\documentclass[twoside,american]{elsarticle}
\usepackage[T1]{fontenc}
\usepackage[utf8]{inputenc}
\usepackage{array}
\usepackage{float}
\usepackage{amsmath}
\usepackage{amssymb}
\usepackage{graphicx}

\makeatletter

\providecommand{\tabularnewline}{\\}
\floatstyle{ruled}
\newfloat{algorithm}{tbp}{loa}
\providecommand{\algorithmname}{Algorithm}
\floatname{algorithm}{\protect\algorithmname}

\journal{arxiv.org}

\usepackage{lmodern}
\usepackage{algorithmicx}
\usepackage{algpseudocode}
\floatname{algorithm}{Function}

\newcommand{\alglinenoNew}[1]{\newcounter{ALG@line@#1}}

\newcommand{\alglinenoPush}[1]{\setcounter{ALG@line@#1}{\value{ALG@line}}}

\usepackage{scrextend}

\makeatother

\usepackage{babel}
\begin{document}
\begin{frontmatter}
\title{Detection of decision-making manipulation in the pairwise comparisons
method}
\author[kis]{Michał Strada}
\ead{mstrada@agh.edu.pl}
\author[kis]{Sebastian Ernst}
\ead{ernst@agh.edu.pl}
\author[msagh]{Jacek Szybowski}
\ead{szybowsk@agh.edu.pl}
\author[kis]{Konrad Kułakowski\corref{cor1}}
\ead{kkulak@agh.edu.pl}
\cortext[cor1]{Corresponding author}
\address[kis]{AGH University of Krakow, Department of Applied Computer Science}
\address[msagh]{AGH University of Krakow, Faculty of Applied Mathematics}
\begin{abstract}
Most decision-making models, including the pairwise comparison method,
assume the decision-makers honesty. However, it is easy to imagine
a situation where a decision-maker tries to manipulate the ranking
results. This paper presents three simple manipulation methods in
the pairwise comparison method. We then try to detect these methods
using appropriately constructed neural networks. Experimental results
accompany the proposed solutions on the generated data, showing a
considerable manipulation detection level.
\end{abstract}
\end{frontmatter}

\section{Introduction}

When it comes to decision-making, especially in high-stakes situations
such as medical diagnoses, financial investments, or political elections,
it is essential to ensure the fairness and transparency of the process
\citep{Bac2001ccat}. Manipulation of the decision-making process
can have serious consequences that can negatively affect individuals
\citep{Kontek2020stoc}, society, or organizations \citep{Gorsira2018cioe}.
Prejudice, external pressures, bribery, or multiple factors can influence
decision-makers, leading to sub-optimal outcomes or harm. In political
elections, propaganda, disinformation, or bribery can manipulate voters
and influence elections \citep{Faliszewski2009lacv}, leading to long-term
societal consequences.

We can take various measures to prevent manipulation and ensure transparency
\citep{Bac2001ccat} and objectivity in decision-making. For example,
the number of decision-makers can be increased to make manipulation
more difficult \citep{Pelta2014atro}, or the decision-making processes
may be subject to external oversight (or review) to ensure compliance
with ethical and legal standards \citep{Bjorn2015gdmi}. Another critical
aspect is data-driven decision-making, where objective and reliable
data support decision-making. By analyzing data, decision-makers can
identify patterns and anticipate and evaluate alternatives more objectively,
reducing the influence of biases and external pressures. 

Researchers have proposed many methods to reduce subjectivity and
increase transparency, ensuring the decision-making process is open,
visible, and not manipulable \citep{Dong2021cras,Kulakowski2024rhao}.
 Many are quantitative methods, such as mathematical models \citep{Brandt2016hocs,Szybowski2024aomo}
or statistical analysis \citep{Ishizaka2021asmd}. For instance, AHP
(Analytic Hierarchy Process) is a widely used method that decomposes
complex decisions into smaller, more manageable components, allowing
decision-makers to evaluate alternatives based on multiple criteria
\citep{Kulakowski2020utahp}.  Using MCDM (multiple-criteria decision-making
methods) can be particularly challenging in contexts where bribery
and corruption are prevalent. Bribery, corruption, or manipulation
can undermine the integrity of the decision-making process \citep{Orgad2019nafc},
leading to decisions based on personal or financial interests rather
than on the merits of the evaluated alternatives. This can have severe
consequences for stakeholders’ welfare and the sustainability of the
decision. 

As a response to this intricate issue, researchers and practitioners
have embraced innovative methodologies to scrutinize, comprehend,
and alleviate the influence of manipulation. This paper delves into
the intersection of AHP, bribery, and the utilization of neural networks,
presenting a comprehensive framework to prevent and combat corrupt
practices. Moreover, in acknowledgment of the constraints of traditional
methods in curbing bribery, there is a burgeoning interest in harnessing
advanced technologies. Neural networks, a tool of machine learning,
a subset of artificial intelligence, have showcased remarkable pattern
recognition and decision-making capabilities. This paper explores
the integration of neural networks as a preventive measure against
bribery, with the aim of creating proactive systems capable of identifying
and deterring corrupt activities before they escalate. 

\section{Preliminaries}

When individuals make decisions, they usually compare possible alternatives
and choose the one that suits them best. For example, when buying
an orange at a store, one may opt for the larger one. When assessing
products on a scale, the comparison involves their weight against
a one-kilogram standard. Comparing alternatives in pairs facilitates
the creation of a ranking system, helping select the best alternative.

In AHP, every alternative is systematically compared against others.
This comparison results in a PC matrix where rows and columns correspond
to alternatives, and individual elements indicate the outcomes of
these comparisons. These matrices serve as the basis for priority
derivation methods, transforming them into weight vectors. The i-th element
of the vector represents the significance of the corresponding i-th alternative.
The reliability of this ranking depends on the consistency of the PC matrix.
There is a consensus that the more consistent the set of comparisons,
the more reliable the assessment becomes. This section will briefly
overview fundamental methods used to compute priorities and estimate
inconsistency for pairwise comparison matrices.

\subsection{PC matrices\protect\label{subsec:PC_matrices}}

In the AHP method, a decision-maker, often an expert in the particular
field, compares the alternatives in pairs. The easiest way to show
these comparisons is in the form of an $n\times n$ square matrix:

\begin{equation}
\boldsymbol{C}=\left[c_{ij}\right],\label{eq:pc_matrix}
\end{equation}

\noindent where $n$ denotes the order of the pairwise comparison
$PC$ matrix (the number of the alternatives), and each $c_{ij}$
element in the matrix $\boldsymbol{C}$ indicates the result of the
comparisons between the alternatives $a_{i}$ and $a_{j}$. The $PC$
matrix $\boldsymbol{C}$ is said to be reciprocal if for every $i,j\in\{1,\ldots,n\}$
the reverse comparison $a_{j}$ to $a_{i}$ gives:

\begin{equation}
c_{ji}=\frac{1}{c_{ij}}.\label{eq:reciprocal}
\end{equation}

There are several different methods allowing us to calculate a ranking
of $n$ alternatives $A=\{a_{1},\ldots,a_{n}\}$ using $\boldsymbol{C}$
as an input. Let us denote the ranking value of $a_{i}$ as $w\left(a_{i}\right)$
and compose the priority vector as:

\begin{equation}
w=\left[w(a_{1}),\ldots,w(a_{n})\right]^{T}.\label{eq:ranking_vector}
\end{equation}

Although there are many methods to calculate the priority vector $w$
\citep{Choo2004acff,Koczkodaj1997aobf,Kou2014acmm}, we will focus
on two of probably the most popular of these \citep{Kulakowski2020utahp,Kulakowski2015otpo}.
The EVM (eigenvalue method) proposed by \citet{Saaty1977asmf} and
the GMM (geometric mean method) proposed by \citet{Crawford1985anot}.

In the EVM \citep{Kulakowski2020utahp}, the priority vector $w$
is calculated a normalized principal eigenvector of $\boldsymbol{C}$
such that\footnote{$\left\Vert \cdot\right\Vert _{m}=1$ is the Manhattan norm.}
$\left\Vert w\right\Vert =1$:

\begin{equation}
w(a_{i})=\frac{\widehat{w}_{i}}{\sum_{j=1}^{n}\widehat{w}_{j}},
\end{equation}

\noindent where $\widehat{w}$ is the principal eigenvector of $\boldsymbol{C}$,
i.e. it satisfies the following equation:

\begin{equation}
\boldsymbol{C}\widehat{w}=\lambda_{\textit{max}}\widehat{w}.\label{eq:EVM_method}
\end{equation}

\noindent where $\lambda_{\textit{max}}$ is the principal eigenvalue.

On the other hand, in the GMM, the priority of the $i^{\textrm{th}}$
alternative is derived as a rescaled geometric mean of the $i^{\textrm{th}}$
row, thus the priority vector $w$ consist of rescaled geometric means
for all the rows of the matrix $\boldsymbol{C}:$

\begin{equation}
w=\varphi\left[\left(\prod_{j=1}^{n}c_{1j}\right)^{\frac{1}{n}},\ldots,\left(\prod_{j=1}^{n}c_{nj}\right)^{\frac{1}{n}}\right]^{T},\label{eq:GMM_method}
\end{equation}

\noindent where $\varphi$ is a scaling factor such that $\left\Vert w\right\Vert =1$.
In the GMM method scaling factor $\varphi=\left(\sum_{i=1}^{n}\left(\prod_{j=1}^{n}c_{ij}\right)^{\frac{1}{n}}\right)^{-1}$.
Both methods have their counterparts for incomplete matrices \citep{Harker1987amoq,kulakowski2020otgm},
i.e. those in which not all comparison values are defined. 

\subsection{Inconsistency\protect\label{subsec:Inconsistency}}

It is commonly assumed that in the optimal scenario, experts comparing
each pair of alternatives assign $c_{ij}$ a value corresponding to
the actual priority ratio between the two alternatives $a_{i}$ and
$a_{j}$. It is therefore natural to expect that

\begin{equation}
c_{ij}=\frac{w\left(a_{i}\right)}{w\left(a_{j}\right)}.\label{eq:Celement_cij_as_ai_div_aj}
\end{equation}

If $c_{ij}=c_{ik}c_{kj}$ holds for all $i,j,k\in\{1,\ldots,n\}$,
the PC matrix $\boldsymbol{C}=\left[c_{ij}\right]$ is said to be
multiplicatively consistent. Otherwise, if this condition is not satisfied,
$\boldsymbol{C}$ is considered as inconsistent.

Each priority vector $w$ induces exactly one consistent PC matrix
$\boldsymbol{C}$ and vice versa, a consistent PC matrix $\boldsymbol{C}$
induces exactly one vector $w$. Thus, for consistent pairwise comparison
matrix, both methods already mentioned, EVM and GMM, result into the
same vector $w$.

Like the PC matrices themselves, their inconsistency is quantitative.
Thus, one may be tempted to quantify to what extent the PC matrix
(decision data in the form of a set of pairwise comparisons of alternatives)
is inconsistent. 

One of the most recognizable and frequently used indices is Saaty's
consistency index $CI$ \citep{Saaty1977asmf} defined as:

\begin{equation}
\textit{CI}\left(\boldsymbol{C}\right)=\frac{\lambda_{\textit{max}}-n}{n-1}.\label{eq:saaty_idx}
\end{equation}

\noindent Additionally, Saaty also introduced the consistency ratio
$CR$ \citep{Saaty1977asmf} defined as:

\begin{equation}
\textit{CR}\left(\boldsymbol{C}\right)=\frac{\textit{CI}(C)}{\textit{RI}(C)},\label{eq:CR_index}
\end{equation}

\noindent where $\textit{RI}(\boldsymbol{C})$ is the average inconsistency
for the fully random $PC$ matrix having the same dimensions as $\boldsymbol{C}$.
As proposed by Saaty every $PC$ matrix $\boldsymbol{C}$ for which
$\textit{CR}\left(\boldsymbol{C}\right)>0.1$ is deemed to be too
inconsistent to form the basis of a ranking. 

Another consistency index used to determine inconsistency is the Geometric
Consistency Index  \citet{Aguaron2003tgci}, defined as:

\begin{equation}
\textit{GCI}\left(\boldsymbol{C}\right)=\frac{2}{(n-2)(n-1)}\sum_{i<j}ln^{2}\left(c_{ij}\frac{w(a_{j})}{w(a_{i})}\right),\label{eq:gci_idx}
\end{equation}

\noindent where $w$ is derived with the use of the GMM method \eqref{eq:GMM_method}.
The threshold value below which the matrix $\boldsymbol{C}$ is considered
sufficiently consistent can be calculated empirically following the
approach proposed by \citep{Saaty1977asmf}.  There are many different
inconsistency indexes for pairwise comparison matrices. There are
at least a dozen different inconsistency indexes for pairwise comparison
matrices including Koczkodaj's index \citep{Koczkodaj2018aoii}, Kazibudzki's
index \citep{Kazibudzki2022oeop}, Brunelli-Cavallo's index \citep{Brunelli2020dbmo},
and others \citep{Brunelli2018aaoi}. The properties of inconsistency
indices are the subject of numerous studies \citep{Brunelli2013apoi,Bozoki2008osak,Csato2019aoii,Kulakowski2019tqoi}.

\subsection{Error measurement\protect\label{subsec:Error_measurement}}

In real-life scenarios, $PC$ matrices created by experts turns out
to be inconsistent. Most often, the value of the inconsistency index
is not sufficient to detect those pairwise comparisons that correspond
to a high level of inconsistency (the Koczkodaj inconsistency index
may be an exception \citep{Koczkodaj2018aoii}). For this reason,
the concept of error is often considered \citep{Saaty2004dmta,Kulakowski2015otpo}. 

To define an error indicator let us consider a $PC$ matrix $\boldsymbol{C},$
where each $c_{ij}$ entry represents an expert's subjective comparison
between alternatives $a_{i}$ and $a_{j}$. The ranking vector $w$,
for $\boldsymbol{C}$ is given as $w=\left[w(a_{1}),\ldots,w(a_{n})\right]^{T}$
\eqref{eq:ranking_vector}, then the error matrix \citep{Saaty2004dmta}
can be defined as:

\begin{equation}
\boldsymbol{E\left(C\right)}=\left[c_{ij}\frac{w(a_{j})}{w(a_{i})}\right],\quad\forall\:i,j\in\{1,\ldots,n\}.\label{eq:error_matrix}
\end{equation}
High error values identify highly inconsistent comparisons and thus
potentially erroneous expert judgements. For the consistent $PC$
matrix $\boldsymbol{C}$, all elements of the error matrix are $1$,
i.e. ($e_{ij}=1)$, where $E(C)=[e_{ij}].$

\subsection{Machine learning\protect\label{subsec:Machine_learning}}

Machine Learning (ML) is a field of Artificial Intelligence (AI) concerned
with algorithms that are able to learn from the offered data, without
being explicitly programmed. The term itself was coined by \citet{Samuel1959ssim}.
The primary branches of ML include \citep{Russel2020aima}:
\begin{itemize}
\item \textit{supervised learning,} where models are trained to predict
the values of output variables (\textit{labels}) based on the input
variables (\textit{features}), by presenting the model with labeled
instances. The function which maps features to labels can be implemented
using a wide spectrum of algorithms,
\item \textit{unsupervised learning}, where the instances are unlabeled,
and the models instead try to learn patterns that can be discovered
in the data, 
\item \textit{reinforcement learning}, where the data is not explicitly
labeled, and the algorithm only receives a performance score to guide
its actions.
\end{itemize}
As $PC$ matrices can be treated as data instances, the main focus
of this paper is on \textit{supervised learning}. One of the most
common challenges associated with this technique is to obtain accurate
predictions\footnote{The pairwise comparison method can also be used for score prediction
\citep{Kulakowski2015hreg,Koczkodaj2017htrt}.} (label values), while maintaining the ability to \textit{generalize}
\citep{2004alom}, i.e. to predict the values of labels for yet unseen
instances or, in other words, to avoid \textit{over-fitting} to the
training data.

The standard approach to supervised learning involves splitting the
data set into the training set and the testing set at a particular
proportion, e.g., $80:20$. The algorithm is trained using the former.
Then, the prediction quality is verified for the latter, which the
algorithm has not yet seen. Various metrics can be used for that purpose,
depending on the task (classification or regression) and the character
of the data.

Both the training set and the testing set should be representative
of the entire dataset. Hence, it is expected to perform shuffling before
the division. However, this can still result in some bias; therefore,
a common approach is to use k-fold cross-validation, where the algorithm
is trained and tested for different splits of the initial dataset
\citep{Fushiki2009eope}.

\noindent Convolutional neural networks (CNNs) represent a foundational
element in deep learning, exhibiting considerable capacity for processing
and comprehending visual data. Inspired by the hierarchical organization
of neurons in the human visual system, CNNs emulate this structure,
enabling them to extract intricate features from images and other
spatial data. Critical Components of CNNs are as follows \citep{Geron2019homl}:
\begin{itemize}
\item \textit{convolutional layers}\textit{\emph{ serve as the foundation
of CNNs, functioning as feature extractors that capture patterns and
structures from raw input data. These layers detect edges, textures,
and other salient features by convolving learnable filters across
the input, enabling the network to discern complex visual patterns.}}
\item \emph{pooling layers} play a pivotal role in spatial downsampling,
reducing the computational burden while retaining essential information.
These layers distill the most salient features through operations
such as max pooling and average pooling, thereby facilitating robustness
and generalization.
\item \emph{activation functions} infuse nonlinearities into the network,
empowering it to learn intricate mappings between input and output.
Activation functions like Rectified Linear Unit (ReLU) enable CNNs
to capture and model intricate relationships within data by introducing
complexity and expressiveness.
\item \emph{fully connected layers} integrate the high-level features extracted
by preceding layers, culminating in decision-making or classification.
These layers, analogous to the brain's association areas, synthesize
abstract representations into actionable insights, guiding the network's
predictions.
\end{itemize}
\noindent Training convolutional neural networks (CNNs) is an iterative
parameter optimization process. The network learns to minimize the
disparity between predicted outputs and ground truth labels during
this process. Through back-propagation and stochastic gradient descent,
CNNs adjust their internal parameters (weights and biases), gradually
converging towards an optimal configuration that maximizes predictive
accuracy.

CNNs can be applied in various solutions, from recognizing handwritten
digits to identifying plant species; convolutional neural networks
(CNNs) excel in assigning labels or categories to images based on
their content, enabling automated analysis and categorization. Empowered
to localize and identify objects within images or videos, CNNs revolutionize
tasks ranging from autonomous driving to surveillance and security,
facilitating real-time analysis and decision-making. By segmenting
images into semantically meaningful regions, CNNs facilitate new avenues
in medical imaging, environmental monitoring, and urban planning,
enabling precise analysis and interpretation. CNNs allow accurate
and reliable facial recognition systems, empowering applications such
as biometric authentication, surveillance, and personalized marketing. 

\section{Problem statement\protect\label{sec:Problem_statement}}

Let us consider a scenario where a decision-maker, an expert in a
particular field, must compare multiple alternatives while facing
external pressures limiting the available time for making judgments.
In a typical situation, the expert creates a $PC$ matrix $\boldsymbol{C}$
(\ref{eq:pc_matrix}) and calculates the ranking vector $w$ (\ref{eq:EVM_method},
\ref{eq:GMM_method}), such that $w(a_{r})>w(a_{p})$, where $r,p\in\{1,\ldots,n\}$.
Assuming that the expert may have a personal interest in promoting
an alternative $a_{p}$ over a reference alternative $a_{r}$, one
possible solution is to create a matrix $\boldsymbol{C}$, keeping
in mind that the alternative $a_{p}$ is better than the $a_{r}$.
In this paper, we focused on an alternative approach -- modifying
the initial $PC$ matrix $\boldsymbol{C}$ to the matrix $\boldsymbol{C}'$
where $w'(a_{r})<w'(a_{p})$. To achieve this, we have developed several
manipulation algorithms based on the observation that increasing $c_{ij}$
generally\footnote{Csató and Petróczy's study revealed that EVM \eqref{eq:EVM_method}
is not monotonic, meaning that increasing $c_{ij}$ does not always
result in an increase in $w(a_{i})$. However, Monte Carlo experiments
showed that this phenomenon is negligible \citep{Csato2020otmo}.
Consequently, in practical scenarios, we can disregard this aspect
for moderately inconsistent $PC$ matrices.} increases $w(a_{i})$ at the expense of all other priority values.

\subsection{Naive algorithm\protect\label{subsec:Naive-algorithm}}

The first manipulation algorithm is the simplest one. Assuming that
the goal is to enhance the alternative $a_{i}$, we will try to increase
all elements in the $i^{\textrm{th}}$ row of $\boldsymbol{C}$.

Let $\alpha$ be a positive real number such that $\alpha>\max\{c_{ij}\}$
where $i,j\in\{1,\ldots,n\}$. In the first step of the algorithm,
we update all elements in the $i^{\textrm{th}}$ row (except $c_{ii}$,
which must be $1$) to $\alpha$ and all elements in the $i^{\textrm{th}}$
column to $1/\alpha$. Without the loss of generality, we may assume
that $i=2$. Thus, the initial form of $\boldsymbol{C}$ and the final
form, denoted by $\boldsymbol{C}'$, are as follows:

\[
\boldsymbol{C}=\begin{bmatrix}1 & c_{12} & \cdots & c_{1n}\\
c_{21} & 1 &  & c_{2n}\\
\vdots &  & \ddots & \vdots\\
c_{n1} & c_{n2} & \cdots & 1
\end{bmatrix}\xrightarrow[\textrm{attack}]{\textsl{}}\begin{bmatrix}1 & 1/\alpha & \cdots & c_{1n}\\
\alpha & 1 &  & \alpha\\
\vdots &  & \ddots & \vdots\\
c_{n1} & 1/\alpha & \cdots & 1
\end{bmatrix}^{'}=\boldsymbol{C}'
\]

Let us present the \textit{naive algorithm} in the form of a structural
pseudocode (code listing \ref{lis:Naive_algorithm,Naive_algorithm}):

\begin{algorithm}[H]
\caption{Naive algorithm}\label{lis:Naive_algorithm,Naive_algorithm}

\begin{algorithmic}[1]

\alglinenoNew{naive_algorithm}

\Function{attack}{$C$, $p$, $r$}\label{Naive_algorithm:function_begin}

\State$C'$ $\leftarrow$ C\label{Naive_algorithm:C'_init_value}

\State$\alpha$ $\leftarrow$ set to the initial value\label{Naive_algorithm:alpha_init_value-1}

\State$indices$ $\leftarrow$ generate indices of elements that
can be modified\label{Naive_algorithm:indices_generation}

\For{each subsequent element $i$ in the $indices$}\label{Naive_algorithm:for_begin}

\State$c'_{pi}$ $\leftarrow$ $\alpha$\label{Naive_algorithm:c'_pi_update}

\State$c'_{ip}$ $\leftarrow$ 1/$c'_{pi}$\label{Naive_algorithm:c'_ip_update}

\EndFor\label{Naive_algorithm:for_end}

\State\Return{$C'$}\label{Naive_algorithm:return_values}

\EndFunction\label{Naive_algorithm:function_end}

\alglinenoPush{naive_algorithm}

\end{algorithmic}
\end{algorithm}

\noindent The \textsc{attack} function of the \textit{naive algorithm
}(line \ref{Naive_algorithm:function_begin}) takes three arguments
as input: the $PC$ matrix $\boldsymbol{C}$ to be modified, the index
of the alternative to be promoted -- $p$ and the index of the reference
alternative -- $r$. The first steps in the algorithm are to create
a local copy of the matrix $\boldsymbol{C}$ (line \ref{Naive_algorithm:C'_init_value})
and set the $\alpha$ factor to a user-defined initial value greater
than 1 (line \ref{Naive_algorithm:alpha_init_value-1}). Next, a collection
of indices is created, except for the index of the $p^{\textrm{th}}$
alternative (line \ref{Naive_algorithm:indices_generation}). Then
the \emph{for} loop is executed (lines \ref{Naive_algorithm:for_begin}-\ref{Naive_algorithm:for_end}).
It iterates through all elements of the $indices$ and performs the
desired update of $c'_{pi}$ and its reciprocal element $c'_{ip}$
(lines \ref{Naive_algorithm:c'_pi_update} and \ref{Naive_algorithm:c'_ip_update}).
The \textsc{attack} function ends by returning the modified matrix
$\boldsymbol{C}'$ (line \ref{Naive_algorithm:return_values}).

The following figure (\ref{fig:Naive's-algorithm-heat-map}) shows
the heat maps for the \textit{naive algorithm}. It is clearly visible
that the $8^{\textrm{th}}$ row and column have been modified, while
the rest of the values remain unchanged.

\begin{figure}[H]
\begin{tabular*}{1\linewidth}{@{\extracolsep{\fill}}>{\centering}m{0.48\linewidth}>{\centering}m{0.48\linewidth}>{\centering}p{0.1\linewidth}}
\includegraphics[width=0.9\linewidth]{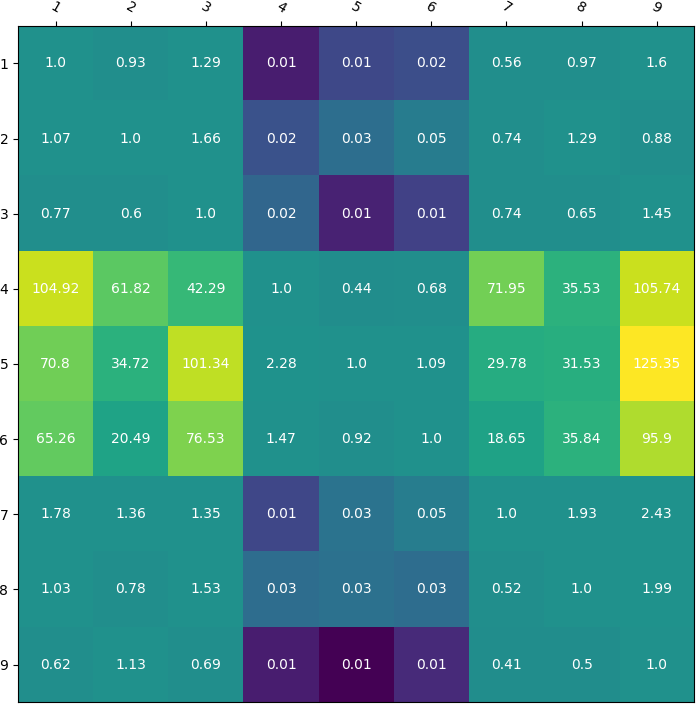} & \includegraphics[width=0.9\linewidth]{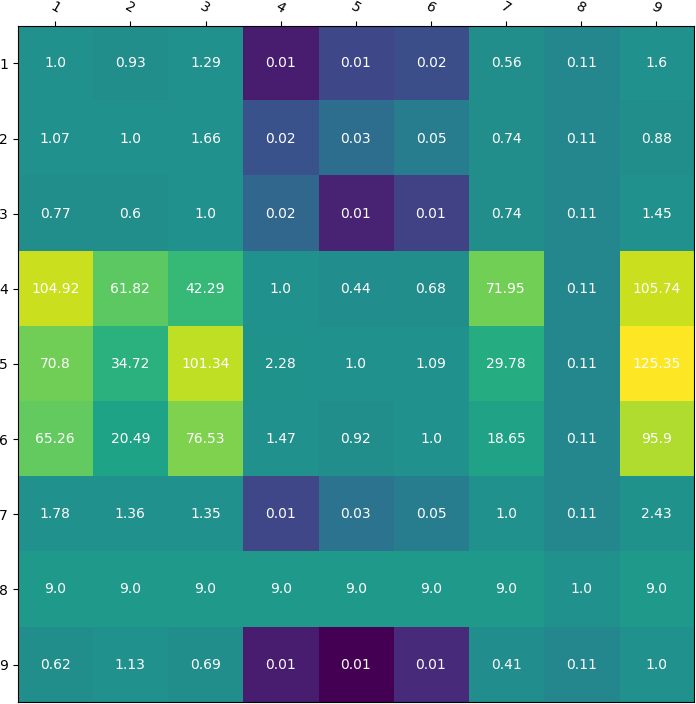} & \tabularnewline
a) heat map of $\boldsymbol{C}$ & b) heat map of $\boldsymbol{C}'$ & \tabularnewline
\end{tabular*}

\caption{Naive algorithm heat map\protect\label{fig:Naive's-algorithm-heat-map}}
\end{figure}

\subsection{Basic algorithm\protect\label{subsec:Basic-algorithm}}

The following manipulation algorithm is an improved version of the
\textit{naive algorithm }\eqref{subsec:Naive-algorithm} and \emph{row
algorithm} \citep{Strada2022manij}, but the main idea behind the
manipulations remains the same. The goal is to improve the ranking
of alternative $a_{i}$ by increasing the values of all elements in
the $i^{\textrm{th}}$ row of $\boldsymbol{C}$.

Let $\alpha$ be a uniformly distributed positive real number over
the right-side open interval $\left[1.1,5\right)$. Similar to the
\textit{naive algorithm}, the first step of the algorithm updates
the value of all elements in the $i^{\textrm{th}}$ row, except for
the $c_{ii}$ element. The value of the $c_{ij}$ is updated by the
$\alpha$ factor multiplied by the corresponding element in the $k^{\textrm{th}}$
row ($c_{kj}$ element). The reciprocity constraint \ref{eq:reciprocal}
is then applied to change all elements in the $i^{\textrm{th}}$ column.
Without the loss of generality, assuming $i=2$ and $k=n$, the initial
form of the $PC$ matrix (denoted by $\boldsymbol{C}$), and the final
form (denoted by $\boldsymbol{C}'$) are as follows:

\[
\boldsymbol{C}=\begin{bmatrix}1 & c_{12} & \cdots & c_{1n}\\
c_{21} & 1 &  & c_{2n}\\
\vdots &  & \ddots & \vdots\\
c_{n1} & c_{n2} & \cdots & 1
\end{bmatrix}\xrightarrow[\textrm{attack}]{\textsl{}}\begin{bmatrix}1 & 1/c_{n1}\alpha & \cdots & c_{1n}\\
\alpha c_{n1} & 1 &  & \alpha\\
\vdots &  & \ddots & \vdots\\
c_{n1} & 1/\alpha & \cdots & 1
\end{bmatrix}^{'}=\boldsymbol{C}'
\]

The following code listing -- \ref{lis:Basic_algorithm,Basic_algorithm},
shows a structural pseudocode of the \textit{basic algorithm}:

\begin{algorithm}[H]
\caption{Basic algorithm}\label{lis:Basic_algorithm,Basic_algorithm}

\begin{algorithmic}[1]

\alglinenoNew{basic_algorithm}

\Function{attack}{$C$, $p$, $r$}\label{Basic_algorithm:function_begin}

\State$C'$ $\leftarrow$ C\label{Basic_algorithm:C'_init_value}

\State$\alpha$ $\leftarrow$ set to the initial value\label{Basic_algorithm:alpha_init_value}

\State$indices$ $\leftarrow$ generate indices of elements that
can be modified\label{Basic_algorithm:indices_generation}

\For{each subsequent element $i$ in the $indices$}\label{Basic_algorithm:for_begin}

\State$c'_{pi}$ $\leftarrow$ $\alpha$$c'_{ri}$\label{Basic_algorithm:c'_pi_update}

\State$c'_{ip}$ $\leftarrow$ 1/$c'_{pi}$\label{Basic_algorithm:c'_ip_update}

\EndFor\label{Basic_algorithm:for_end}

\State\Return{$C'$}\label{Basic_algorithm:return_values}

\EndFunction\label{Basic_algorithm:function_end}

\alglinenoPush{basic_algorithm}

\end{algorithmic}
\end{algorithm}

\noindent The initial steps of the \textsc{attack} function for the
\textit{basic algorithm }(lines \ref{Basic_algorithm:function_begin}-\ref{Basic_algorithm:indices_generation})
are similar to those in the \textit{naive algorithm} (code listing
\ref{lis:Naive_algorithm,Naive_algorithm}, lines \ref{Naive_algorithm:function_begin}-\ref{Naive_algorithm:indices_generation}).
However, selecting the value of the $\alpha$ factor (line \ref{Basic_algorithm:alpha_init_value}),
follows a different approach, as it is uniformly chosen from a predefined
range. Next, the for loop is executed (lines \ref{Basic_algorithm:for_begin}-\ref{Basic_algorithm:for_end}).
The code iterates through all elements of the $indices$ and updates
$c'_{pi}$ (line \ref{Basic_algorithm:c'_pi_update}) and its reciprocal
element $c'_{ip}$ (line \ref{Basic_algorithm:c'_ip_update}) as desired.
The last step of the \textsc{attack} function returns the modified
matrix $\boldsymbol{C}'$ (line \ref{Basic_algorithm:return_values}).

The following figure (\ref{fig:Basic's-algorithm-heat-map}) shows
the heat maps for the \textit{basic algorithm}. The values in the
$8^{\textrm{th}}$ row have been updated with the values from the
$9^{\textrm{th}}$ row multiplied by the $\alpha$ factor (and similarly
all reciprocal elements), while the rest of the values remain unchanged.

\begin{figure}[H]
\begin{tabular*}{1\linewidth}{@{\extracolsep{\fill}}>{\centering}m{0.48\linewidth}>{\centering}m{0.48\linewidth}>{\centering}p{0.1\linewidth}}
\includegraphics[width=0.9\linewidth]{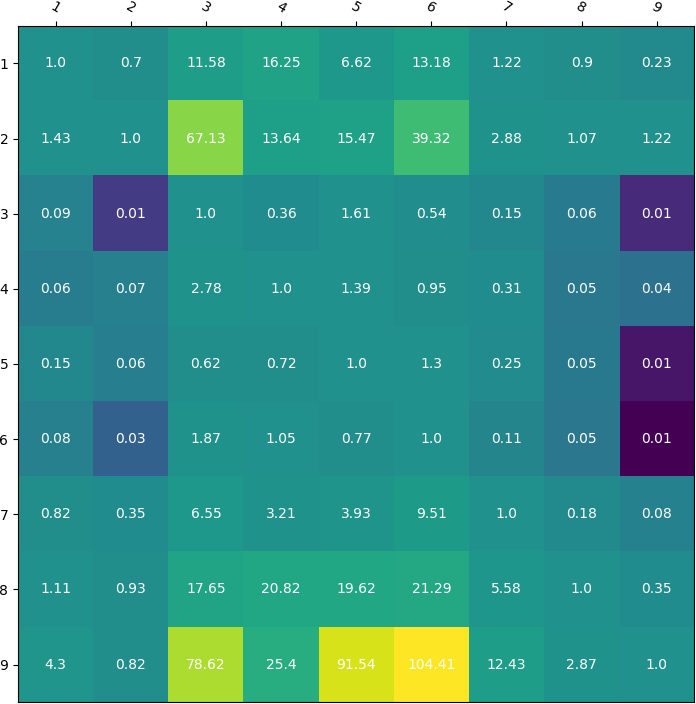} & \includegraphics[width=0.9\linewidth]{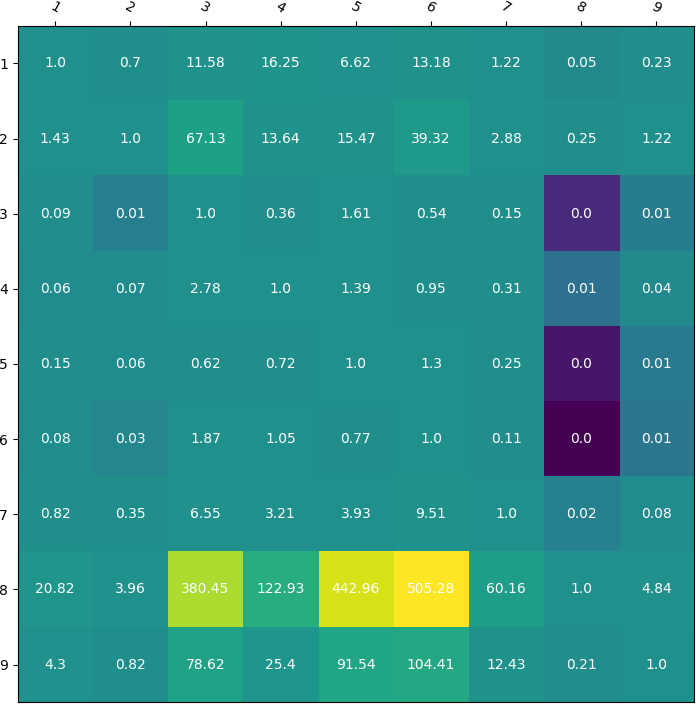} & \tabularnewline
a) heat map of $\boldsymbol{C}$ & b) heat map of $\boldsymbol{C}'$ & \tabularnewline
\end{tabular*}

\caption{Basic algorithm heat map\protect\label{fig:Basic's-algorithm-heat-map}}
\end{figure}

\subsection{Advanced algorithm\protect\label{subsec:Advanced-algorithm}}

This chapter describes the \textit{advanced algorithm}, which is essentially
the \textit{basic algorithm }\eqref{subsec:Basic-algorithm} with
a stopping mechanism. The goal is to improve the ranking of alternative
$a_{i}$ by increasing the values of as few elements as possible in
the $i^{\textrm{th}}$ row and column of the matrix $\boldsymbol{C}$. 

The $\alpha$ factor in this algorithm has the same properties as
in the \textit{basic algorithm}. The $c_{ij}$ element in the $i^{\textrm{th}}$
row and its reciprocal counterpart $c_{ji}$ in the $i^{\textrm{th}}$
column are updated sequentially, except for the $c_{ii}$ element,
as long as the ranking value \eqref{eq:ranking_vector} of the alternative
$a_{i}$ is lower than the ranking value of the reference alternative
$a_{k}$ or all feasible elements are updated. The updated value of
the $c_{ij}$ element is equal to the corresponding $c_{kj}$ element
multiplied by the $\alpha$ factor. Without the loss of generality,
we may assume that $i=2$ and $k=n$. Thus, the initial form of $\boldsymbol{C}$,
the intermediate form denoted by $\boldsymbol{C}^{\textrm{step}},$
and the final form denoted by $\boldsymbol{C}'$, are as follows:

\[
\begin{array}{c}
\boldsymbol{\quad\hspace{0.88em}C}=\begin{bmatrix}1 & c_{12} & \cdots & c_{1n}\\
c_{21} & 1 &  & c_{2n}\\
\vdots &  & \ddots & \vdots\\
c_{n1} & c_{n2} & \cdots & 1
\end{bmatrix}\xrightarrow[\textrm{first step}]{\textsl{}}\begin{bmatrix}1 & 1/c_{n1}\alpha & \cdots & c_{1n}\\
\alpha c_{n1} & 1 &  & c_{2n}\\
\vdots &  & \ddots & \vdots\\
c_{n1} & c_{n2} & \cdots & 1
\end{bmatrix}^{'}=\boldsymbol{C}^{1}\\
\\\downarrow\textrm{next }n-2\textrm{ consecutive steps}
\end{array}
\]

\[
\boldsymbol{C}^{n-2}=\begin{bmatrix}1 & c_{12} & \cdots & c_{1n}\\
c_{21} & 1 &  & c_{2n}\\
\vdots &  & \ddots & \vdots\\
c_{n1} & c_{n2} & \cdots & 1
\end{bmatrix}\xrightarrow[\textrm{last \hspace{0.35em}step}]{\textsl{}}\begin{bmatrix}1 & 1/c_{n1}\alpha & \cdots & c_{1n}\\
\alpha c_{n1} & 1 &  & \alpha\\
\vdots &  & \ddots & \vdots\\
c_{n1} & 1/\alpha & \cdots & 1
\end{bmatrix}^{'}=\boldsymbol{C}'
\]

\noindent It is important to note that when the stopping criteria
are met, the number of algorithm steps may be lower than those ($n-1$
steps) shown in the example above. The \textit{advanced algorithm}
as a structural pseudocode (code listing \ref{lis:Advanced_algorithm,Advanced_algorithm}):

\begin{algorithm}[H]
\caption{Advanced algorithm}\label{lis:Advanced_algorithm,Advanced_algorithm}

\begin{algorithmic}[1]

\alglinenoNew{advanced_algorithm}

\Function{attack}{$C$, $p$, $r$}\label{Advanced_algorithm:function_begin}

\State$C'$ $\leftarrow$ C\label{Advanced_algorithm:C'_init_value}

\State$\alpha$ $\leftarrow$ set to the initial value\label{Advanced_algorithm:alpha_init_value}

\State$indices$ $\leftarrow$ generate indices of elements that
can be modified\label{Advanced_algorithm:indices_generation}

\For{each subsequent element $i$ in the $indices$}\label{Advanced_algorithm:for_begin}

\State$c'_{pi}$ $\leftarrow$ $\alpha$$c'_{ri}$\label{Advanced_algorithm:c'_pi_update}

\State$c'_{ip}$ $\leftarrow$ 1/$c'_{pi}$\label{Advanced_algorithm:c'_ip_update}

\If{$w(a'_{p})$ > $w(a'_{r})$}\label{Advanced_algorithm:break_condition}

\State\textbf{break}\label{Advanced_algorithm:break_point}

\EndIf\label{Advanced_algorithm:break_condition_end}

\EndFor\label{Advanced_algorithm:for_end}

\State\Return{$C'$}\label{Advanced_algorithm:return_values}

\EndFunction\label{Advanced_algorithm:function_end}

\alglinenoPush{advanced_algorithm}

\end{algorithmic}
\end{algorithm}

\noindent The first steps for the \textit{advanced algorithm }(lines
\ref{Advanced_algorithm:function_begin}-\ref{Advanced_algorithm:indices_generation})
are identical to those of the \textit{basic algorithm }(code listing
\ref{lis:Basic_algorithm,Basic_algorithm}, lines \ref{Basic_algorithm:function_begin}-\ref{Basic_algorithm:indices_generation}).
Next, the for loop (lines \ref{Advanced_algorithm:for_begin}-\ref{Advanced_algorithm:for_end})
iterates through all elements of the $indices$ and updates element
$c'_{pi}$ and its reciprocal counterpart element $c'_{ip}$ (line
\ref{Advanced_algorithm:c'_pi_update} and \ref{Advanced_algorithm:c'_ip_update}).
Then, the stopping criterion is evaluated (line \ref{Advanced_algorithm:break_condition}).
If the ranking of the alternative to be promoted ($a_{p}$) is greater
and the ranking of the reference alternative ($a_{r}$) the for loop
is exited (line \ref{Advanced_algorithm:break_point}), otherwise,
iteration through the next elements of the $indices$ continues. The
final step of the \textsc{attack} function returns the modified matrix
$\boldsymbol{C}'$ (line \ref{Advanced_algorithm:return_values}).

The following figure (\ref{fig:Advanced's-algorithm-heat-map}) shows
the heat maps for the \textit{advanced algorithm}. Some values in
the $8^{\textrm{th}}$ row have been updated with the values from
the $2^{\textrm{nd}}$ row multiplied by the $\alpha$ factor (and
similarly all reciprocal elements), while the rest remain unchanged.

\begin{figure}[H]
\begin{tabular*}{1\linewidth}{@{\extracolsep{\fill}}>{\centering}m{0.48\linewidth}>{\centering}m{0.48\linewidth}>{\centering}p{0.1\linewidth}}
\includegraphics[width=0.9\linewidth]{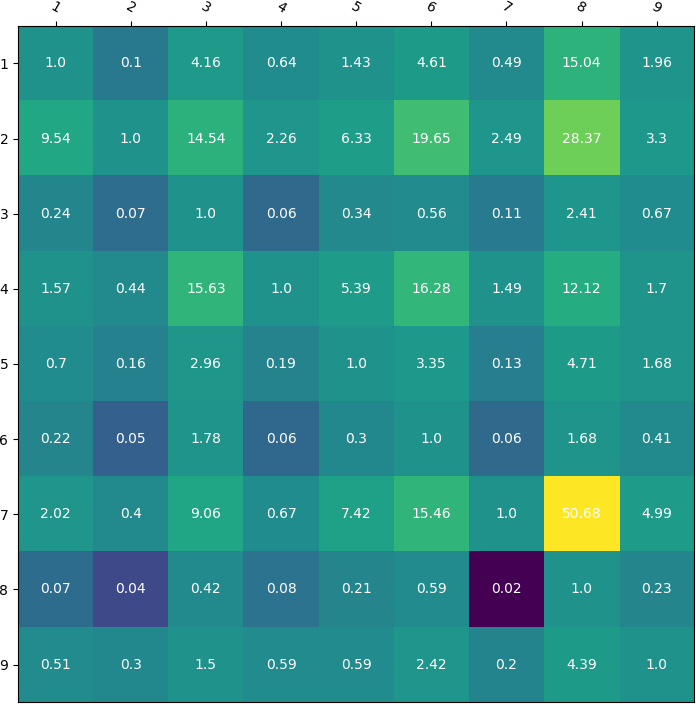} & \includegraphics[width=0.9\linewidth]{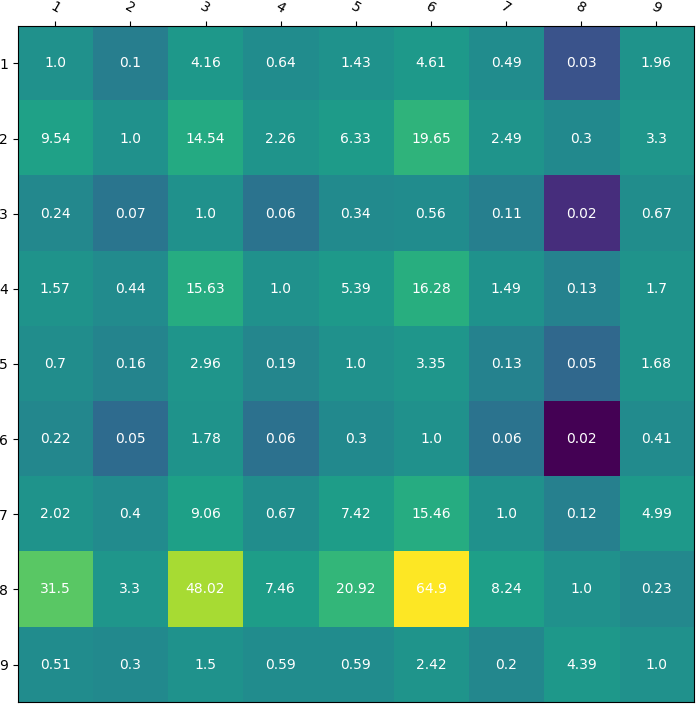} & \tabularnewline
a) heat map of $\boldsymbol{C}$ & b) heat map of $\boldsymbol{C}'$ & \tabularnewline
\end{tabular*}

\caption{Advanced algorithm heat map\protect\label{fig:Advanced's-algorithm-heat-map}}
\end{figure}

\section{Detecting manipulation using Machine Learning\protect\label{sec:Detecting-manipulation-using-ML}}

As mentioned in the subsection \ref{subsec:Machine_learning}, the
primary focus of the presented work is on \textit{supervised learning}
methods. One of the most important tasks when building such models
is to choose the algorithm which is most appropriate for the characteristics
of the data in question.

$PC$ matrices are two-dimensional objects. Therefore, they are similar
to raster images, such as the images of handwritten digits present
in the MNIST dataset \citep{LiDeng2012tmdo}. They can be seeded as
2-D images, representing the comparison values, e.g., as different
shades of gray.

\begin{figure}[H]
\begin{centering}
\includegraphics[width=0.8\linewidth]{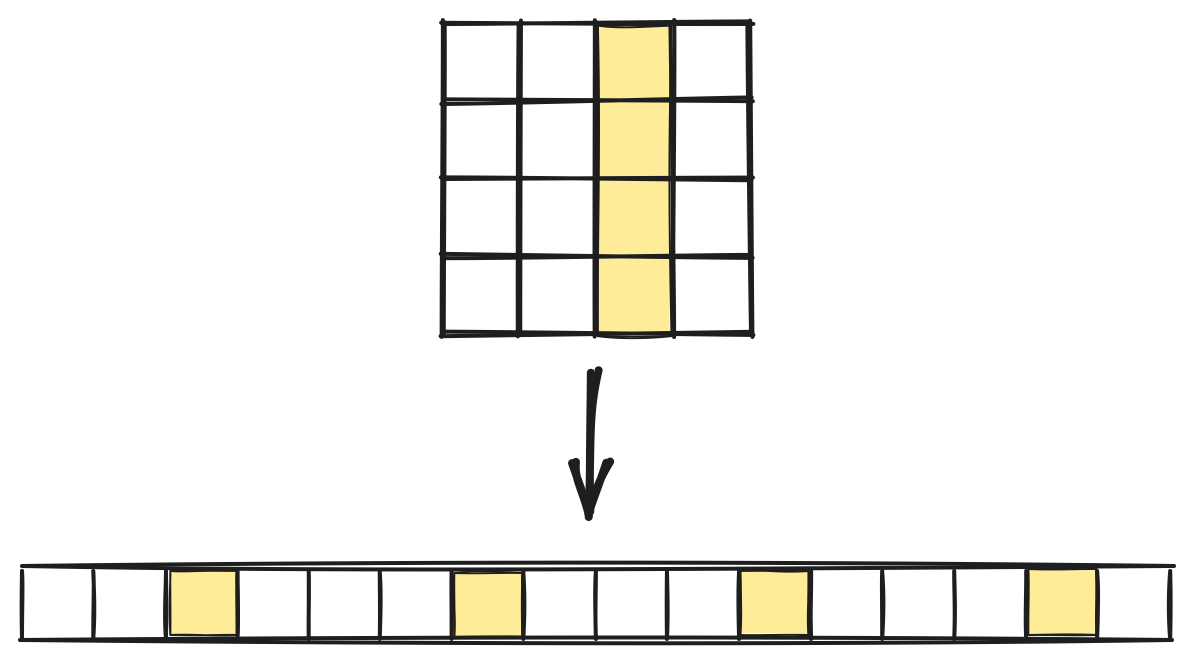}
\par\end{centering}
\caption{A simple (2-D) $PC$ matrix being flattened to a (1-D) vector.\protect\label{fig:matrix-flatten}}
\end{figure}

However, for most ML algorithms, the input object is transformed to
a vector. This may result in the algorithm not being able to “see”
certain phenomena in the data. For instance, if the \textit{naive}
manipulation consists in maximizing the values in a single matrix
column, visually forming a vertical line, this fact may not be apparent
after the data had been flattened, as presented in Fig. \ref{fig:matrix-flatten}.

Therefore, it seems reasonable to explore ML algorithms that can detect
2-D relationships between values. One of the most prominent classes
in this category is convolutional neural networks (CNNs) \citep{Geron2019homl}.
 Compared to fully connected neural networks, these significantly
reduce the number of weights (parameters) by utilizing convolution
kernels (also called filters).

The following sections explore the potential of using CNNs to detect
manipulations in PC matrices, describe the tuning process of the network
structure, and present the results of the experiments. 

\section{Experiments and results\protect\label{sec:Experiments_and_results}}

To test the neural network's accuracy in detecting the use of the
described algorithms, we first conducted a training process and then
a series of experiments. To detect the use of each of the three attack
algorithms, a separate network has been prepared. We repeated the
neural network training process on $PC$ matrix $\boldsymbol{C}$
of different sizes $N$ .

To ensure that the results of the described algorithms are comparable,
we have chosen the following common input parameters:
\begin{itemize}
\item $\boldsymbol{\alpha}$ -- for the \textit{naive algorithm} it is
equal to the size of the $PC$ matrix $\boldsymbol{C}$, otherwise
it is a uniformly distributed random number between $1.1$ and $5$,
\item $\boldsymbol{r}$ -- index of the reference alternative, always equal
to the index of the highest ranked alternative evaluated from the
initial matrix $\boldsymbol{C}$,
\item $\boldsymbol{p}$ -- index of the alternative to be promoted, randomly
selected and different from $\boldsymbol{r}$.
\end{itemize}
All the matrices used in the experiments were generated as follows:
\begin{itemize}
\item first, a ranking vector $w$ \eqref{eq:ranking_vector} of length
$n$ was generated,
\item next, the ranking vector $w$ was transformed into a consistent $PC$
matrix $\boldsymbol{C}$ using \eqref{eq:Celement_cij_as_ai_div_aj},
\item finally, the matrix $\boldsymbol{C}$ was disturbed to ensure that
its inconsistency \eqref{eq:CR_index} was greater than $0$.
\end{itemize}
We generated and attacked $10\,000$ $PC$ matrices for each $N\in\{5,6,\ldots,9\}$
and for each tested manipulation algorithm. A total of $300\,000$
samples of different sizes were generated and manipulated. Next, matrices
were divided into two sets: one set, containing $20\%$ of the samples,
was used to evaluate the model, and the other set, containing $80\%$
of the samples, was used to train the model.

The attack detection rate $\left[\%\right]$ is shown in the table
below:

\begin{table}[H]
\begin{centering}
\begin{tabular*}{1\linewidth}{@{\extracolsep{\fill}}|>{\centering}m{0.1\columnwidth}|>{\centering}m{0.25\columnwidth}|>{\centering}m{0.25\columnwidth}|>{\centering}m{0.25\columnwidth}|}
\hline 
$N$ & \textit{naive algorithm} & \textit{basic algorithm} & \textit{advanced algorithm}\tabularnewline
\hline 
$5$ & $89$ & $99$ & $93$\tabularnewline
\hline 
$6$ & $96$ & $99$ & $93$\tabularnewline
\hline 
$7$ & $98$ & $\sim100$ & $92$\tabularnewline
\hline 
$8$ & $99$ & $\sim100$ & $92$\tabularnewline
\hline 
$9$ & $\sim100$ & $99$ & $90$\tabularnewline
\hline 
\end{tabular*}
\par\end{centering}
\caption{Attack detection rate\protect\label{tab:Attack-detection-rate-3DConv}}
\end{table}

To evaluate the model, we used pre-processed sample matrices. Our
initial attempt to train a neural network on bare matrices $\boldsymbol{C}$
and $\boldsymbol{C}'$ yielded unacceptable results. Next, we attempted
to use the error matrix \eqref{eq:error_matrix} in the pre-processing
step to achieve better results. However, the detection rate for the
\textit{basic algorithm }oscillated around $65\%$, which is also
unacceptable.

Finally, in the pre-processing step, we have used the following $\mathbb{R}^{2}\rightarrow\mathbb{R}^{3}$
transformation:

\begin{equation}
\boldsymbol{D}\left(\boldsymbol{C}\right)=\left[d_{i,j,k}\right],\quad d_{i,j,k}=\det\left[\begin{array}{ccc}
c_{i,i} & c_{i,j} & c_{i,k}\\
c_{j,i} & c_{j,j} & c_{j,k}\\
c_{k,i} & c_{k,j} & c_{k,k}
\end{array}\right],\quad\forall\:i,j,k\in\{1,\ldots,n\}\label{eq:determinant-factor}
\end{equation}

The following figure (\ref{fig:design-of-nn}) shows the design of
a neural network model for different values of $N$:

\begin{figure}[H]
\begin{tabular*}{1\columnwidth}{@{\extracolsep{\fill}}>{\centering}m{0.48\columnwidth}>{\centering}m{0.48\columnwidth}}
\includegraphics[width=0.48\columnwidth]{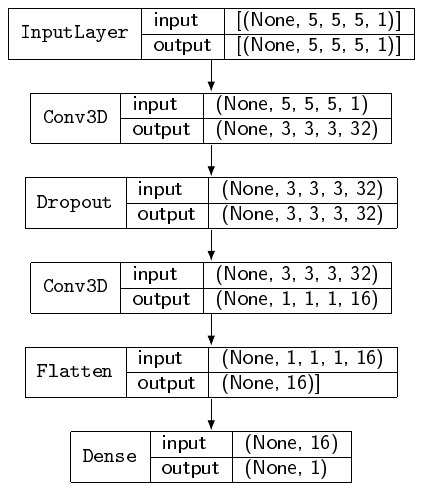} & \includegraphics[width=0.48\columnwidth]{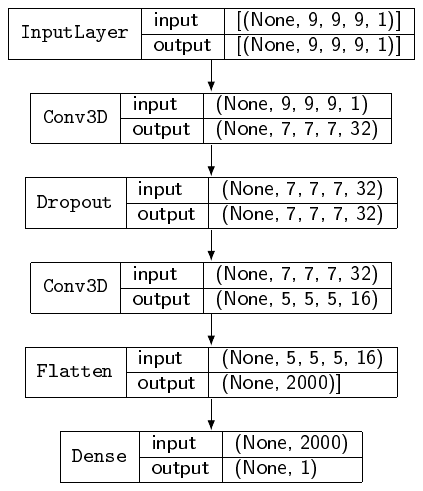}\tabularnewline
a) $N=5$\label{fig:Neural-network-model-design-a} & b) $N=9$\label{fig:Neural-network-model-design-b}\tabularnewline
\end{tabular*}

\caption{Design of Neural Networks used in the experiments\protect\label{fig:Neural-network-model-design}}
\label{fig:design-of-nn}
\end{figure}

\noindent It is worth noting that the network's design is simple.
However, the detection rate is high, reaching $100\%$ in some cases.
For $N=5$, the Flatten layer is unnecessary and is only included
for consistency within the model.

\section{Discussion\protect\label{sec:Discussion-and-conclusions}}

The experiments performed and then described in section \ref{sec:Experiments_and_results}
show that designed algorithms effectively manipulate the selected
alternatives' rankings. As previously stated, simple indicators such
as the error matrix \eqref{eq:error_matrix} or inconsistency indices
\eqref{eq:CR_index} and \eqref{eq:gci_idx} are insufficient for
adequately training the neural network. We think some features of
the manipulation algorithms are only detectable when more than two
ranking values are compared simultaneously. Based on this observation,
we introduced the determinant factor \eqref{eq:determinant-factor}.
We have also investigated other methods for transforming $\mathbb{R}^{2}$
$PC$ matrices to $\mathbb{R}^{3}$ factors. However, we are still
researching their impact on detecting attacks by neural networks.  

Our experiments used 2-D and 3-D convolutional layers with the same
number of hidden neurons and layers for both cases. We achieved the
most optimal results for 3-D models, although models with only one
hidden layer can also produce acceptable results. The detection rates
for the 2-D network, which has a similar design as shown in \ref{fig:Neural-network-model-design},
with the error matrix as a pre-processing step, are as follows:

\begin{table}[H]
\begin{centering}
\begin{tabular*}{1\linewidth}{@{\extracolsep{\fill}}|>{\centering}m{0.1\columnwidth}|>{\centering}m{0.25\columnwidth}|>{\centering}m{0.25\columnwidth}|>{\centering}m{0.25\columnwidth}|}
\hline 
$N$ & \textit{naive algorithm} & \textit{basic algorithm} & \textit{advanced algorithm}\tabularnewline
\hline 
$5$ & $89$ & $86$ & $79$\tabularnewline
\hline 
$6$ & $95$ & $77$ & $80$\tabularnewline
\hline 
$7$ & $98$ & $68$ & $82$\tabularnewline
\hline 
$8$ & $99$ & $67$ & $83$\tabularnewline
\hline 
$9$ & $\sim100$ & $58$ & $82$\tabularnewline
\hline 
\end{tabular*}
\par\end{centering}
\caption{Attack detection rate (2-D approach).\protect\label{tab:Attack-detection-rate-2dConv}}
\end{table}

\noindent We used \eqref{eq:determinant-factor} in the preprocessing
step because the detection rate of the basic algorithm falls below
our threshold.

\section{Summary\protect\label{sec:Summary}}

Unfortunately, decision-making methods based on quantitative pairwise
comparisons are prone to manipulation. In this work, we proposed a
manipulation model and three attack algorithms. We also analyzed the
effectiveness of the neural networks in detecting attacks for different
sizes of pairwise comparison matrices. At the preprocessing stage,
we considered various methods to ensure the optimal data format processed
in subsequent phases. Detecting manipulation is a significant challenge
for researchers. In future research, we plan to investigate this issue
further, particularly in the context of different and more complex
manipulations. Further investigation is required into the preprocessing
step for more advanced manipulation techniques, and the neural network
models used to detect them. Therefore, this matter will be a challenge
for us in future research.

\section*{Acknowledgements}

The research has been supported by the National Science Centre, Poland,
as a part of the project SODA no. 2021/41/B/HS4/03475. Jacek Szybowski
has also been partially supported by the Polish Ministry of Science
and Higher Education within the internal task of AGH University of
Krakow no. 11.11.420.004. This research was funded in whole or in
part by National Science Centre, Poland 2021/41/B/HS4/03475. For the
purpose of Open Access, the author has applied a CC-BY public copyright
license to any Author Accepted Manuscript (AAM) version arising from
this submission.

\bibliographystyle{plain}
\bibliography{__from_DOI,papers_biblio_reviewed,bibliography_reviewed_ms}

\end{document}